\documentclass{article}
\usepackage{graphicx} 
\usepackage{authblk}
\usepackage{hyperref}
\usepackage{amssymb} 
\usepackage{booktabs}
\usepackage{multirow}
\usepackage{array}
\usepackage{algorithm}
\usepackage{algorithmic}
\usepackage{amsmath}
\usepackage{float}
\usepackage{rotating}
\usepackage{siunitx}

\title{Explicit Grammar–Semantic Feature Fusion for Robust Cross-Domain Text Classification}
\author[1]{Azrin Sultana (25-93678-1@student.aiub.edu)}
\author[1, *]{Firoz Ahmed (fahmed@aiub.edu)}
\affil[1]{Department of Computer Science, American International University-Bangladesh, Dhaka-1229, Bangladesh}
\affil[*]{Corresponding author: fahmed@aiub.edu}

\begin{document}

\maketitle
\section{Abstract}
Natural Language Processing (NLP) enables computers to understand human language and analyze and classify text efficiently by processing deep-level grammatical and semantic features. Existing models capture features either by learning from large corpora with transformer models, which are computationally intensive and unsuitable for resource-constrained environments, or by using lightweight shallow models. Previous studies overlooked the importance of incorporating comprehensive grammatical rules alongside semantic information to build a robust, lightweight classification model without resorting to full-parameterized transformer models or heavy deep learning architectures. To address this gap, we proposed a novel Grammar-Aware Feature Engineering framework for cross-domain text analysis. The novelty of our approach lies in its explicit encoding of sentence-level grammatical structure—including syntactic composition, phrase patterns, and complexity indicators—into a compact grammar vector, which is then fused with frozen contextual embeddings. These heterogeneous elements unified a single representation that captures both the structural and semantic characteristics of the text. Deep learning models such as Deep Belief Network (DBN), Long Short-Term Memory (LSTM), BiLSTM, and transformer-based BERT and XLNET were used to train and evaluate the model, with the number of epochs varied. Based on experimental results, the unified feature representation model captures both the semantic and structural properties of text, outperforming baseline models by 2\%-15\%, enabling more effective learning across heterogeneous domains.  Unlike prior syntax-aware transformer models that inject grammatical structure through additional attention layers, tree encoders, or full fine-tuning, the proposed framework treats grammar as an explicit inductive bias rather than a learnable module resulting in a very lightweight model that delivers better performance on edge devices.

\section{Introduction}

Natural Language Processing (NLP) processes textual data, which enables computers to understand and communicate in human language. It enables machines to recognize, understand, and generate text and speech by combining rule-based modelling of human language with statistical and machine learning models. \cite{i1} \cite{i2}. The demand for the NLP market is increasing significantly, projected to approximately 15.1\% from 2023 to 2030 \cite{i3}. Over the past 20 years, computational linguistics has grown into both a scientific field and a practical technology. This growth is attributed to the increasing demand for NLP solutions across various industries and applications, including healthcare, finance, e-commerce, and customer service \cite{i4}. 

NLP tasks primarily involve syntactic and semantic analysis of the text \cite{i5}. It is used for a wide range of language-related tasks, including answering questions, classifying text, and conversing with users. Its typical applications include sentiment analysis of text, primarily reviews; spam detection in email analysis; and token classification, also referred to as named entity recognition (NER), which labels each word in a sentence \cite{i6}. Question answering of an automatic chatbot system for product queries, where the model examines the question's context and identifies the relevant information in the passage to provide an accurate answer \cite{i7}. Currently, NLP is widely used for translation generation from one language to another \cite{i8} \cite{i9}. The earliest applications of NLP were simple if-then decisions that required preprogrammed rules to perform the NLP task. They can provide answers only in response to specific prompts because rule-based NLP lacks machine-learning or AI capabilities. Statistical NLP automatically extracts, classifies, and labels elements of text and assigns a statistical likelihood to each possible meaning of those elements. This relies on machine learning to enable sophisticated analysis of linguistic phenomena \cite{i10}. Statistical NLP introduced the essential technique of mapping language elements, such as words and grammatical rules, to vector representations, enabling language modeling using mathematical methods, including regression and Markov models \cite{i12}. 

Model performance largely depends on the quality of data used during training. Feature engineering is a crucial preprocessing technique for NLP tasks, as machines cannot process raw text. Feature representation transforms raw text into a structured representation that a machine can use to extract meaningful information \cite{i14} \cite{i15}. Traditional NLP feature engineering techniques, such as bag-of-words, treat an unordered collection of words, disregarding word order and most syntactic or grammatical information \cite{i16} \cite{i17}. Another statistical method is TF-IDF, which quantifies how important a word is to a document relative to a larger corpus \cite{i18}. With advances in deep learning, semantic feature representations such as Word2Vec \cite{i19}, GloVe \cite{i20}, and FastText \cite{i21} embeddings have become widely used. Given a sequence of words in a sentence, the Word2Vec model takes a fixed number of context words as input, and each is represented as an embedding via a shared embedding layer. In contrast, GloVe generates dense vector representations by analyzing their co-occurrence patterns in a corpus. FastText can capture the semantic meaning of morphologically related words, even for out-of-vocabulary words or rare words, making it particularly useful for handling languages with rich morphology or for tasks where out-of-vocabulary words are common \cite{i21}. Among these feature selection methods, traditional approaches ignore word order, grammar, and context and are sensitive to domain-specific vocabulary. Transformer-based models leverage dense contextual embeddings to achieve state-of-the-art performance across a variety of tasks. However, while these models capture semantic and syntactic relationships, they often do so implicitly within their embedding space and are less robust across domains with divergent linguistic structures \cite{i22}. Additionally, to obtain semantic and synthetic information from text, transformer models need to be trained with parameter counts ranging from millions to hundreds of billions. Training and even inference require massive matrix multiplications, which are energy-intensive on GPUs/TPUs, making it difficult to run these heavy models on resource-limited devices. Most syntax-aware transformer models incorporate syntax through additional attention heads, tree encoders, or fine-tuning, which increases parameter count and training cost. This contradicts deployment requirements for low-resource and edge environments \cite{i13}. Therefore, critical grammatical signals—such as explicit syntactic dependencies, phrase-level patterns, and sentence complexity—may be underutilized or obscured during downstream learning.

The core limitation of existing syntax-aware transformer architectures is that grammatical structure is introduced through additional trainable components, increasing model capacity and computational cost while tightly coupling syntax learning to domain-specific lexical statistics. In contrast, this work proposes a fundamentally different paradigm: grammar is introduced as an explicit inductive bias that constrains representation learning without modifying or fine-tuning the transformer architecture. By freezing the transformer backbone and injecting a low-dimensional grammar vector at the representation level, the proposed framework restricts the hypothesis space toward linguistically plausible configurations, improving domain robustness while preserving model efficiency. Based on these limitations, we proposed a novel, extensive, a grammar-based feature selection technique, which integrates a rich set of grammar-driven rules to understand the text features in depth with syntactic and structural insights, including POS, syntactic dependency relations, phrase structure patterns, verb-adverb ratio, structure of the sentence, and complex sentence indicators to capture the underlying grammatical style and discourse flow of each document. To incorporate contextual information, the extracted sentence-level grammar features are fused with token-level embeddings, thereby giving each token access to the global sentence context.  Through this hybrid feature design, the model learns domain-invariant grammatical regularities, improving both interpretability and robustness in cross-domain classification tasks.

The main contribution of our proposed model:
\begin{itemize}
  \item Designing a grammar-based feature vector with text embedding for a classification task
  \item Integrated 11 different grammar rules for extracting grammar-based features.
  \item extensive experiments on both document-level classification and token-level NER tasks, showing consistent performance gains across deep learning and transformer-based models while maintaining a lightweight architecture.
  \item Evaluate and compare the effectiveness of deep learning  LSTM, BiLSTM, DBN, and transformer-based models, including  BERT, DistilBERT, using these combined grammar and contextual feature vectors.
  \item To assess model performance in both binary and multi-class settings, we used accuracy, precision, recall, and F1-score.
\end{itemize}

\section{Literature Review}
The purpose of text understanding is to enable machines to accurately interpret human language. Therefore, it is necessary to represent text as efficient feature vectors using extensive grammatical and transformer-based learning so that the machine can learn complex patterns. Recent research on grammar-based feature engineering for NLP tasks remains limited. 

Existing work, Mohasseb et al.\cite{r1}, explored a grammar-based \cite{r1}, explored a grammar-based question-answer classification framework. Three experimental settings were considered: baseline classification, class-balanced classification via SMOTE, and binary classification. On the contrary, in \cite{r2}, a new firefly algorithm-based feature selection method is proposed. To validate this technique, a Support Vector Machine classifier is used, along with three evaluation metrics: precision, recall, and F1-score. Furthermore, experiments on the OSAC real dataset, along with comparisons with state-of-the-art methods, are conducted. POS tagging has been widely used in deep learning to extract features from text; recent research is modifying POS models with new algorithms to make the system more robust and less error-prone. \cite{r3} used Markov models and the Viterbi algorithm with POS. In contrast, Han et al. \cite{r4} improved the POS tagging model for imperative sentences, which differ in syntax, and trained the POS tagger on a new corpus. With this technique, tagging precision increased by 27\%. The lexical ambiguity is resolved by extracting local contextual knowledge from attached images to help users better understand the process. POS tagging is also used for Thai wh-question classification in text, employing feature selection and word embedding techniques, as in Chotirat et al. \cite{r5}. conducted several experiments to evaluate the performance of the proposed methodology using two different datasets (the TREC-6 dataset and the Thai sentence dataset), including Unigram, Unigram+Bigram, and Unigram+Trigram. POS tagging, with rule-based data preprocessing proposed by  Li et al. \cite{r6}. By masking a portion of the POS tags and using self-attention, the model can leverage bidirectional context. The proposed approach combines rule-based methods with deep learning, which is beneficial for POS tagging research. Experiments on the GMB dataset for POS tagging achieved a whole-sentence accuracy of 76.04\%.Kou et al. \cite{r7} proposed a model with a multi-criteria decision system-based evaluation method to assess feature selection methods' performance for classification tasks with small numbers of samples. 
Feature selection methods such as bag-of-words and TF-IDF with transformer models are a new research direction. The authors use this to train the model to learn feature embeddings with greater accuracy. Chaudhary et al. \cite{r8} proposed an architecture with TF-IDF feature extraction and GPT-2 integration, yielding a 4.6\% improvement in precision and recall. A binary sentiment classification of user reviews in Brazilian Portuguese is proposed in \cite{r9} that provides a comprehensive experimental study of embedding approaches. This study includes classical Bag-of-Words to state-of-the-art Transformer-based models, evaluated with five open-source databases. 
To understand short texts, Li et al. \cite{r10} proposed a combined method that combines knowledge-based conceptualization with a transformer encoder. For each term in a brief text, obtain its concept base based on cooccurrence terms and concepts, construct a convolutional neural network to capture local context information, and introduce the subnetwork structure based on a transformer embedding encoder and embed these into a low-dimensional vector space to obtain more attention from these concepts based on a transformer with a precision of 81\%.

In summary, current grammar-based feature selection for textual data approaches have demonstrated potential but remain fragmented, rule-bound, and domain-specific. In \cite{r1},\cite{r2} applied a question answering model; however, their models are limited to predefined grammar rules based on question answers without considering the contextual information. Whereas the Viterbi algorithm, applied to sentiment analysis using a hidden Markov model \cite{r3}, is limited to a single dataset. There is a need to develop a generalizable grammar-aware feature engineering framework that can adapt to multiple domains and languages. Our proposed study can advance the field by unifying grammatical multiple rules, complex structures such as phrasal verbs, which are overlooked by almost all studies, into a learnable grammar embedding layer, applying cross-domain classification experiments, mainly sentiment, topic modeling, and question classification, and introducing grammar-aware balancing and dimensionality reduction techniques with transformer embedding. Additionally, our proposed system integrates text embeddings with grammatical features via concatenation, enabling the model to jointly capture syntactic structure and semantic representation for more effective training and representation learning. To the best of our knowledge, this is the first approach to combine in-depth grammatical structures with contextual text embeddings into a unified feature representation, resulting in a more informative and robust modeling framework. 
\section{Methodology}
This chapter describes the process adopted to design and experiment with the proposed grammar-based feature identification. 
\subsection{Dataset description}

\begin{figure}[tbp]
\centering
\includegraphics[width=\textwidth]{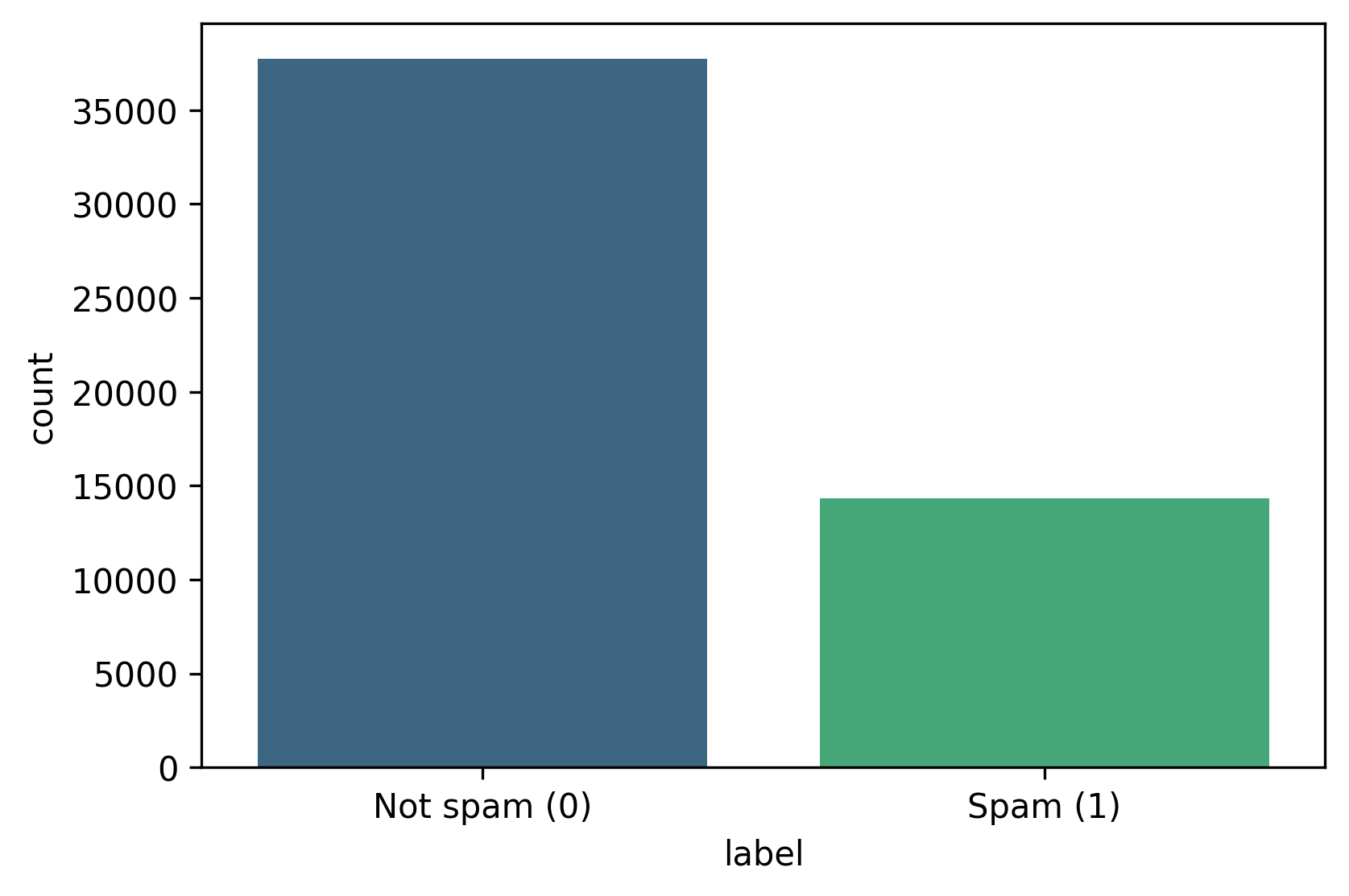}
\caption{Classification dataset class distribution}\label{email_class_distribution}
\end{figure}

\begin{figure}[tbp]
\centering
\includegraphics[width=\textwidth]{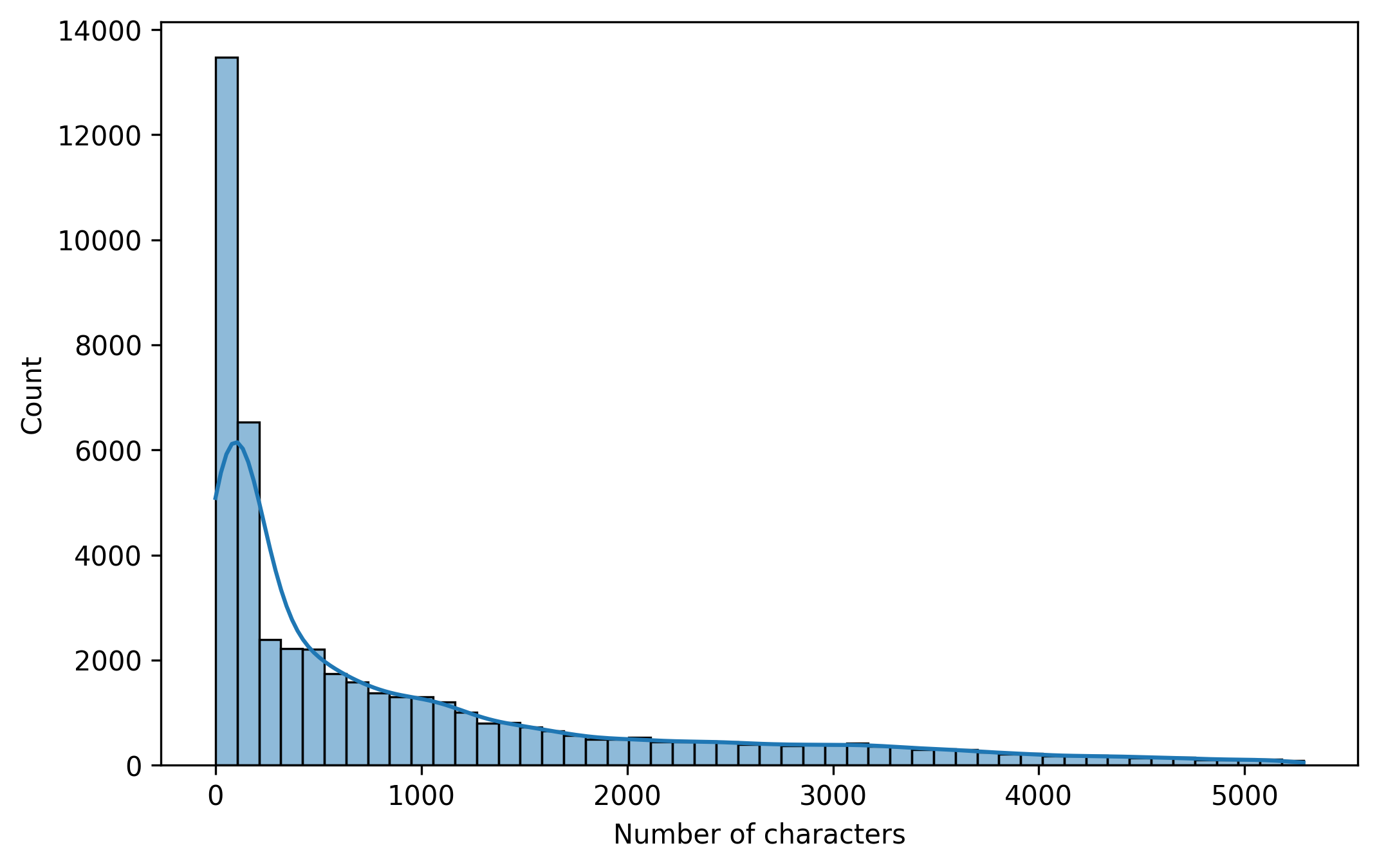}
\caption{Classification dataset text data length distribution}\label{email_length_distribution}
\end{figure}

\begin{figure}[tbp]
\centering
\includegraphics[width=\textwidth]{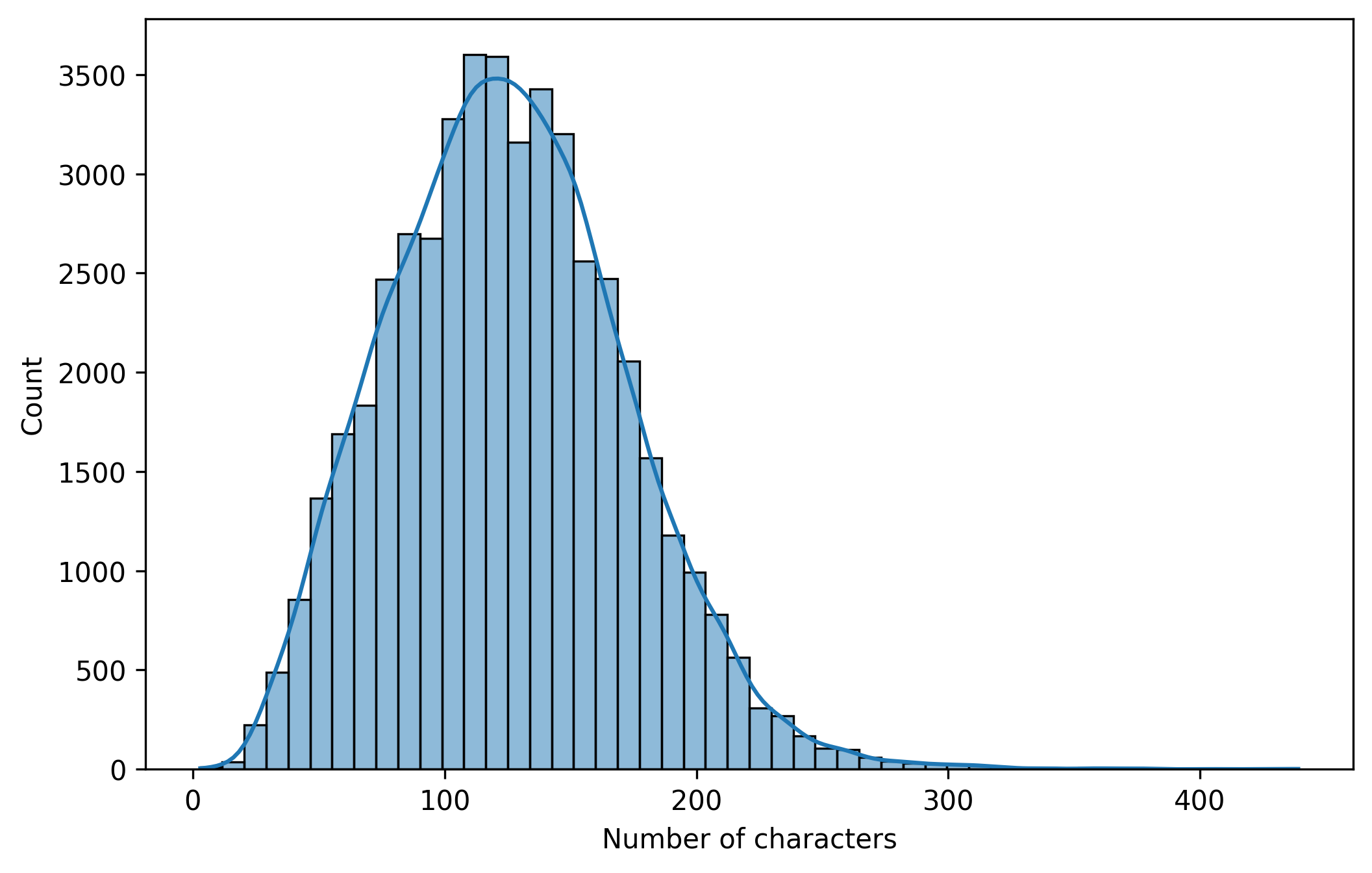}
\caption{NER dataset sentence length distribution}\label{ner_length_distribution}
\end{figure}

For this study, we used two domain-specific datasets to evaluate the effectiveness of the proposed classification model and to capture deeper linguistic information using grammar-based features. Using two datasets allows us to assess the generality and robustness of grammar-based features across diverse linguistic settings and natural language processing (NLP) tasks, including classification, sentiment analysis, and named entity recognition. For the classification task, the email classification dataset \cite{m2} is used. The classification dataset has 52,062 instances. Initially, the dataset contains multiple columns; however, for our task, we retained only the text and label attributes. The dataset consisted of 52,062 messages, of which 72.47\% (37729) were categorized as “not spam” and 14,333 (27.53\%) were labeled as "spam", as depicted in Figure \ref{email_class_distribution}. From the figure \ref{email_length_distribution} approximately 14,000 email lengths are close to zero, depicting that most of the emails in this dataset are short; however, there are some email lengths that are above 5000 characters. For the NER model, the widely famous GNB dataset is used \cite{m1}. Each token in this dataset is assigned a label indicating entity (e.g., person, organization, location) or a miscellaneous category. Labels are given in table \ref{ner_labels}. The dataset is extracted from the GMB corpus, which is tagged and explicitly annotated for training a classifier to predict named entities, such as names and locations. GMB is a relatively large corpus of 1 million data points with extensive annotations. It is labelled using the IOB tagging system, in which each entity label is prefixed with either B or I. B denotes the beginning, and I inside of an entity. The words that are not of interest are labelled with 0. Sentence number, word, and POS are the features available in the dataset. The tag column, which is our target column, assigns a tag to each word in a sentence. From the count plot below, the distribution of the sentence character can be observed in Figure \ref{ner_length_distribution}. From the figure, most of the sentence lengths are between 100 and 200 characters.

\begin{table}[tbp]
\centering
\caption{Named Entity Categories Used in the GMB Dataset}
\begin{tabular}{ll}
\hline
\textbf{Label} & \textbf{Description} \\
\hline
GEO  & Geographical Entity \\
ORG  & Organization \\
PER  & Person \\
GPE  & Geopolitical Entity \\
TIM  & Time Indicator \\
ART  & Artifact \\
EVE  & Event \\
NAT  & Natural Phenomenon \\
\hline
\label{ner_labels}
\end{tabular}
\end{table}

\subsection{Grammar as an Explicit Inductive Bias for Lightweight and Domain-Robust Learning}

Although transformer-based architectures have the theoretical potential to model syntactic dependencies via self-attention, these syntactic regularities are not explicitly inducted; instead, they are implicitly inferred from data. The quality and generality of the learned syntactic representations remain, therefore, highly dependent on the lexical composition and domain characteristics of the training corpus. When exposed to limited, noisy, or stylistically divergent data, transformer models tend to internalize domain-specific lexical correlations rather than domain-invariant structural patterns, resulting in poor generalization performance.

In contrast, features induced from grammar capture linguistic regularities that are, for the most part, domain-independent and independent of surface vocabulary. They cover aspects of structural properties, such as phrase hierarchy, dependency relations, and compositionality constraints, which remain stable across genres, topics, and styles. By incorporating such grammar-based informant signals into the representation space, the current framework makes an explicit inductive bias toward linguistically plausible configurations. This biases the hypothesis space toward structural coherence, reducing reliance on domain-specific terminology and overfitting under small or heterogeneous training conditions.

Importantly, it is emphasized that this proposed approach does not try to discard or suppress any kind of representation learning by transformers. Nor does it need to retrain or re-task complex syntax-conscious transformer models. It should be noted that the grammar feature provides structural cues that supplement semantic representations. This enables a desired balance between structural guidance and expressive semantic capacity within a single model. This serves as a very valuable alternative to end-to-end approaches that combine transformers with syntax.

Furthermore, with regard to grammar-based features, it can be argued that, in line with a learning theory perspective, grammar-based features should be understood as having provided an inductive bias in their capacity as analogues of a prior applied over the space of possible hypotheses. Finally, the strong consistency in model performance improvement across diverse data sets and model configurations supports the view that grammar-based inductive biases provide unconditional benefits for both model architecture and domain-specific semantics. Across diverse data sets and model configurations support the view that grammar-based inductive biases provide unconditional benefits with regard to both model architecture and domain-specific semantics.

\subsection{Model diagram}
\begin{figure}[htbp]
\centering
\includegraphics[width=\textwidth]{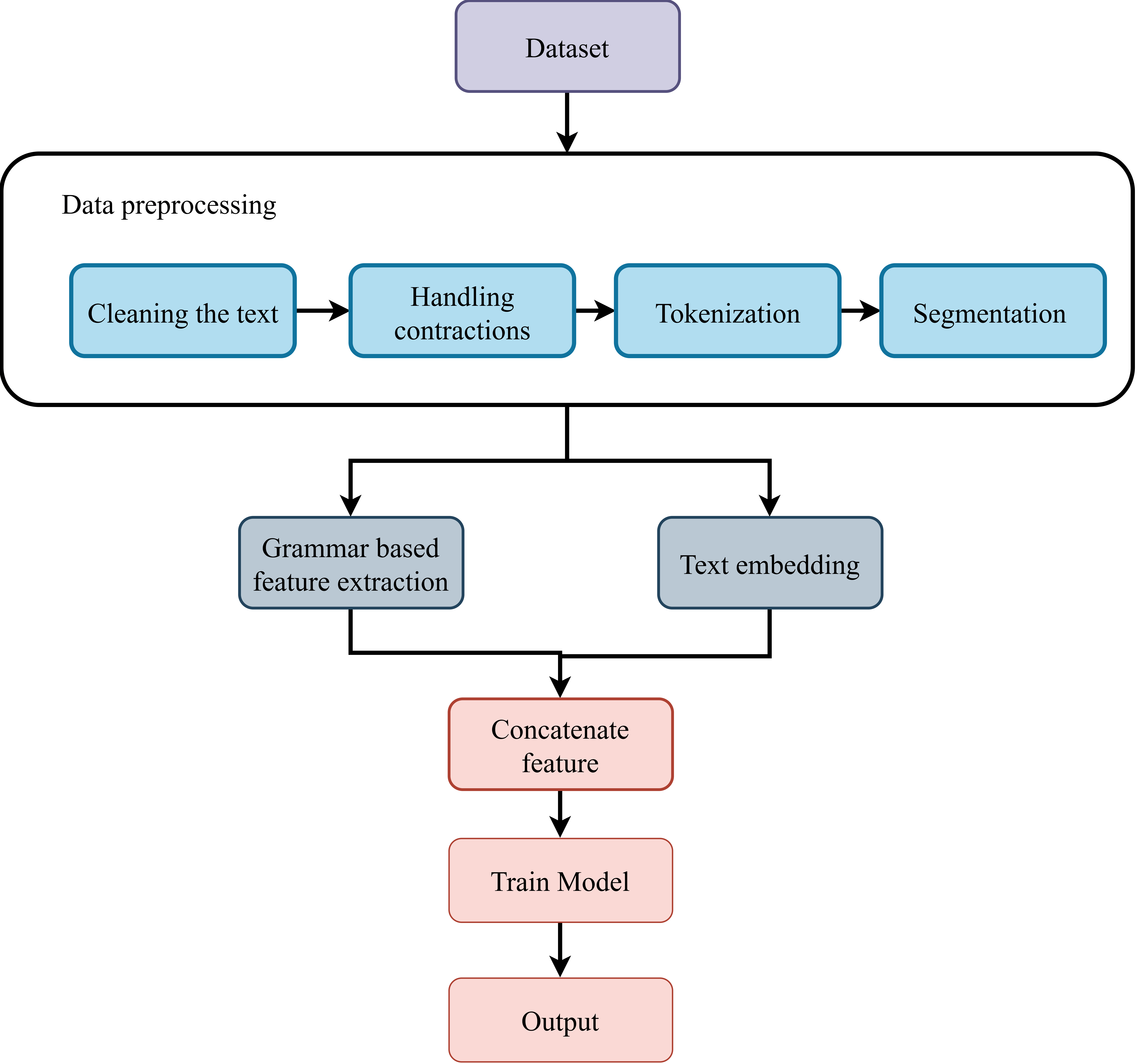}
\caption{Model diagram of the proposed system}\label{model}
\end{figure}
We proposed a hybrid, grammar-aware, transformer-based learning framework to improve text classification performance across multiple domains. The Figure \ref{model} shows the model architecture of our proposed system, comprising four steps: dataset preprocessing, extraction of grammar- and text-based features, feature fusion, and model training and evaluation. 
\subsubsection{Text processing} 
In Natural Language Processing (NLP), the manipulation, analysis, and interpretation of textual data using computational methods are crucial for enabling machines to understand and analyze natural language \cite{m13}. Text preprocessing is a vital step in preparing raw text data for NLP models. It involves cleaning and transforming text to ensure that the algorithms can understand and train the model effectively. Because both datasets contain emojis, emoticons, spaces, and punctuation, they may complicate the robust training of our models. Therefore, during the preprocessing and cleaning phase, we removed emoticons, emojis, and extra spaces, and then corrected spelling. In the email classification dataset, many words were incorrect. The cleaned text is then fed into two parallel feature-extraction paths \cite{m14}.

\subsubsection{Feature extraction}
Feature extraction is a fundamental task that converts raw text data into a format that is easily processed by machine learning algorithms \cite{m15}. After preprocessing the text, grammar-based and context features were extracted. Grammar-based features include syntactic and structural insights, such as POS, syntactic dependency relations, phrase-structure patterns, verb-adverb ratio, clause count, parse tree, and complex-sentence indicators, to capture the underlying grammatical style and discourse flow of each document. These depth features in text help models to identify complex patterns in the text. the vector is constructed from grammatical and textual features and then merged with the vector representation of the sentence. 

\subsubsection{Model training and validating}
To measure the effectiveness of the proposed model, we have used multiple deep learning and transformer-based models. We split the dataset 80:20 for training and testing the model.   

\subsection{Proposed grammar feature extraction model}
 Feature extraction is a technique in NLP that converts text into vector representations, reducing dimensionality and data complexity to improve algorithmic performance and efficiency. The Advanced grammar feature extractor is designed to encode linguistic and syntactic properties of text into numerical features. 
\subsubsection{Building grammar-based feature vector}
Syntactic analysis \cite{m12} gets the structure or grammar of the sentences. Processing a sentence syntactically involves determining the subject and predicate and the place of nouns, verbs, pronouns, etc.   Table \ref{tab:feature_groups_detailed} provides the details of grammar-based features. POS tagging's primary role is to help NLP systems understand the syntactic structure of text, which is critical for higher-level tasks such as parsing, semantic analysis, and machine translation. Calculating the POS ratio in a sentence would help detect sentence composition and writing style that vary by task \cite{m16}. Now for a text sentence \(S\) containing \[
S = \{t_1, t_2, \dots, t_N\}
\] tokens, and the set of POS tokens \(P\) for our proposed model are \[
P = \{\text{NOUN},\ \text{VERB},\ \text{ADJ},\ \text{ADV},\ \text{PRON},\ \text{PROPN},\ \text{ADP},\ \text{DET},\ \text{NUM},\ \text{CCONJ}\}
\]
\begin{equation}
\text{POS}_p(S) = \frac{\sum_{i=1}^{N} \mathbf{1}_{\{\text{POS}(t_i) = p\}}}{\sum_{i=1}^{N} \mathbf{1}_{\{\text{POS}(t_i) \in P\}}}
\end{equation}
Sentence Length is an important metric for assessing Readability level, Writer's style, and Text complexity. It also shows how complex the sentence is, as the longer the sentence, the more difficult it is to interpret. For a sentence \(S\) with \(t\) tokens, the average sentence length formula is

\begin{equation}
Avg\_length = \frac{1}{N} \sum_{i=1}^{N} |S_i|
\end{equation}
Here, N is the number of sentences in the text.
Function words establish relationships between words, showing tense, possession, or connecting ideas. Determiners, conjunctions, auxiliaries, and articles are function words. These include words such as he, the, and those, as well as and, or, and but. The more function words present in the sentence, the more complicated the structure becomes, whereas missing function words denote that the structure is incorrect. Based on this feature, our model can also distinguish between formal and informal text, as academic texts contain more function words \cite{m17}. From a predefined function words set \(F\), our model detects those, and the formula is  

\begin{equation}
    R_F = \frac{| \{ t_i : t_i \in F \} |}{T}
\end{equation}

Here, T is the number of total tokens in the text.

A noun phrase is a group of two or more words headed by a noun that includes modifiers. A typical noun phrase consists of a noun together with zero or more dependents of various types. Noun phrases typically bear argument functions \cite{m18}. Every sentence is structured around a verb and a predicate, and every predicate is considered as a verb phrase; .  It represents a crucial technique in NLP for identifying and isolating the core semantic units within a sentence. In information extraction models, such as NER models that incorporate these noun and verb phrases, systems can efficiently identify entities, events, and relations. These two noun and verb phrase formulas count how many phrases are in the text.
\begin{equation}
\text{NP\_count} = 
\sum_{k=1}^{K} 
\mathbf{1}\!\left(
\text{tokens}_{k} \in \{\text{NOUN, PROPN, ADJ}\}^{+}
\right)
\end{equation}

\begin{equation}
\text{VP\_count} =
\sum_{i=1}^{N}
\mathbf{1}\!\left(
POS(t_i) \in V
\;\land\;
\left( i = 1 \;\lor\; POS(t_{i-1}) \notin V \right)
\right)
\end{equation}
Here, \(V = \{\text{AUX}, \text{VERB}, \text{ADV}, \text{PART}\}\)

A clause is a combination of words that make up a sentence. It consists of a subject and a predicate. There are two types of clauses: independent and dependent.  The clause count used to build a grammar-based feature vector directly measures syntactic maturity, cognitive complexity, and expressive capacity in texts \cite{m19}. Our clause-count formula determines the number of clausal relations in the dependency tree.

\begin{equation}
\text{clause\_count} =
\sum_{i=1}^{N}
\mathbf{1}\big( \text{dep}(t_i) \in C \big)
\end{equation}

\[
C = \{\text{csubj},\, \text{csubjpass},\, \text{ccomp},\, \text{advcl},\, \text{acl},\, \text{xcomp}\}
\]
The ratio of adverbs to verbs indicates to the model whether the sentence is descriptive or complex. The larger the ratio value, the greater the emphasis on the manner, time, or frequency of actions. 
\begin{equation}
\text{adv\_verb\_ratio} = \frac{N_{\text{ADV}}}{N_{\text{VERB}} + \varepsilon}
\end{equation}

\noindent
where
\begin{align*}
N_{\text{ADV}} & = \text{number of adverbs in the sentence or text} \\
N_{\text{VERB}} & = \text{number of verbs (may include auxiliaries if desired)} \\
\varepsilon & = \text{small constant (e.g., } 10^{-6} \text{) to avoid division by zero}
\end{align*}
The passive voice represents that the subject is one acted upon by the action or verb in the sentence. In our proposed model, we calculated the passive ratio to assess the formal or descriptive writing style.

\begin{equation}
\text{passive\_ratio} = \frac{N_{\text{passive}}}{N_{\text{VERB}} + \varepsilon}
\end{equation}

\noindent
where
\begin{align*}
N_{\text{passive}} & = \text{number of verbs in passive constructions} \\
N_{\text{VERB}} & = \text{total number of verbs} \\
\varepsilon & = \text{small constant (e.g., } 10^{-6} \text{) to avoid division by zero}
\end{align*}

A parse tree is an ordered tree that represents the syntactic structure of a string or a sentence. It helps identify the sentence's complexity \cite{m20}. The parse tree equation based on our proposed model
\begin{equation}
\text{parse\_tree\_depth} = \frac{1}{N} \sum_{i=1}^{N} d(t_i)
\end{equation}

\noindent
where
\begin{align*}
d(t_i) & = \text{depth of token } t_i \text{ in the dependency parse tree} \\
N & = \text{total number of tokens in the sentence or text}
\end{align*}

When all the features of your grammar-based model are ready, we have created a fixed-length feature vector of 
\[
F = 
\begin{bmatrix}
\text{POS\_ratio} & \text{Avg\_length} & R_f \\
\text{NP\_count} & \text{VP\_count} & \text{Clause\_count} \\
\text{adv\_verb\_ratio} & \text{passive\_ratio} & \text{parse\_tree\_depth}
\end{bmatrix}
\]

\begin{table}[tbp]
\centering
\caption{Linguistic Feature Groups with Descriptions and Advantages}
\begin{tabular}{|l|p{5cm}|p{5cm}|}
\toprule
\textbf{Feature Group} & \textbf{Description} & \textbf{Advantage} \\
\midrule
POS Ratios & Ratio of different part-of-speech tags (nouns, verbs, adjectives, etc.) & Captures syntactic distribution and word-class patterns \\
\hline
Sentence Length & Number of words or tokens per sentence & Indicates structural complexity \\
\hline
Function Words & Frequency of prepositions, conjunctions, articles, pronouns & Reflects grammatical correctness \\
\hline
NP/VP Counts & Count of noun phrases and verb phrases & Captures phrase structure patterns \\
\hline
Clause Count & Number of clauses (main and subordinate) per sentence & Measures syntactic complexity \\
\hline
Adv/Verb Ratio & Ratio of adverbs to verbs & Stylistic and linguistic emphasis \\
\hline
Passive Ratio & Proportion of passive voice constructions & Formality indicator \\
\hline
Parse Tree Depth & Maximum depth of syntactic parse tree & Overall sentence complexity \\
\bottomrule
\end{tabular}

\label{tab:feature_groups_detailed}
\end{table}

\begin{algorithm}[H]
\caption{ Grammar Feature Extractor}
\begin{algorithmic}[1]
\STATE \textbf{Class GrammarExtractor:}
\STATE $\triangleright$ INIT():
\STATE \quad Load spaCy, add sentencizer, define $F$, $P$
\STATE
\STATE $\triangleright$ count\_clauses($doc$):
\STATE \quad $C \gets \{\text{csubj},\dots,\text{xcomp}\}$
\STATE \quad \textbf{return} $\#\{t\in doc \mid t.dep\in C\}$
\STATE
\STATE $\triangleright$ parse\_depth($doc$):
\STATE \quad \textbf{for} $t\in doc$: $d\gets 1+\text{steps\_to\_root}(t)$
\STATE \quad \textbf{return} $\text{mean}(depths)$ or 0
\STATE
\STATE $\triangleright$ passive\_ratio($doc$):
\STATE \quad $V\gets\{t\in doc\mid t.pos=\text{VERB}\}$
\STATE \quad $P_v\gets\{v\in V\mid v.tag=\text{VBN}\lor v.dep=\text{auxpass}\}$
\STATE \quad \textbf{return} $|P_v|/|V|$ if $|V|>0$ else 0
\STATE
\STATE $\triangleright$ extract($texts$):
\STATE \quad \textbf{for} $doc$ in nlp.pipe($texts$):
\STATE \quad \quad \textsf{1. POS ratios:} $R_p = \frac{\#\{t\mid t.pos=p\}}{\sum_{p\in P}\#\{t\mid t.pos=p\}}$
\STATE \quad \quad \textsf{2. Avg sent len:} $L = \text{mean}(|s|$ for $s\in doc.sents)$
\STATE \quad \quad \textsf{3. Func word ratio:} $F_r = \frac{\#\{t\in F\}}{|doc|}$
\STATE \quad \quad \textsf{4. NP/VP counts:} Count sequences, count verbs
\STATE \quad \quad \textsf{5. Advanced:} clauses, adv/verb, passive, parse depth
\STATE \quad \textbf{return} $\text{DataFrame}\to\text{numpy array}$
\end{algorithmic}
\end{algorithm}

\subsubsection{Fusion of Grammar-Based Features with Contextual Embeddings}
To enrich token-level representations with explicit syntactic information, we integrate the extracted grammar features with contextual embeddings derived from a pre-trained BERT model. In semantic analysis, word sense disambiguation is the automated process of determining a word's sense in a given context. As natural language consists of words with several meanings, the objective here is to recognize the correct meaning based on its use. Semantic analysis facilitates the processing of queries and their interpretation, thereby enabling an organization to discern the customer’s preferences.  The goal of semantic analysis is to extract exact meaning, or dictionary meaning, from the text. Extracting both the semantic and syntactic structures of the sentence would provide a more precise representation for learning complex patterns from texts.


Let a sentence \(S = \{t_1, t_2, \dots, t_N\}\)
have a grammar feature vector \(G(S) \in \mathbb{R}^{d_g}\) and BERT produce contextual embeddings 
\[
E = [e_1, e_2, \dots, e_N] \in \mathbb{R}^{N \times d_b}
\] 
Where $d_g$ and $d_b$ are the dimensions of the grammar vector and BERT embeddings, respectively.

Concatenation can be applied:
\begin{equation}
    \tilde{e}_i = [\, e_i \, \Vert \, G'(S) ]\
\end{equation}
where $\Vert$ denotes vector concatenation. The fused embeddings 
\begin{equation}
    \tilde{E} = [\tilde{e}_1, \dots, \tilde{e}_N]
\end{equation}
are then fed into downstream layers, such as a transformer encoder or a task-specific classifier.

Now, the fuse vector combines contextual information about the text with linguistic details about its structure.





\subsection{Algorithm used}
\subsection{Deep Belief Network}
A DBN is a graphical representation of a deep neural network composed of multiple layers of hidden variables, with connections between layers but not within each layer. It can be denoted as a collection of simple learning models, each a restricted type of RBM. 
A DBN is constructed by stacking multiple RBMs trained in a greedy layer-wise fashion.
The joint distribution is defined as:
\begin{equation}
P(\mathbf{v}, \mathbf{h}) = \frac{1}{Z} \exp\left(-E(\mathbf{v}, \mathbf{h})\right)
\end{equation}
Here \(\mathbf{h}^{(1)},......,\mathbf{h}^{(L)}\) denotes hidden layers
DBN uses a loss function \(\mathcal{L}_{\text{CE}}\) to measure the error between the predicted and true value. 
\begin{equation}
    \mathcal{L}_{\text{CE}} = -\frac{1}{N} \sum_{n=1}^{N} \sum_{i=1}^{C} y_{n,i} \log(\hat{y}_{n,i})
\end{equation}
here, \(N\) is the number of samples, \(y_{n,i}\) true level, and \( \log(\hat{y}_{n,i})\) error in predicted level.

A layer of hidden units represents features that capture higher-order patterns in the text's correlations. The key idea behind DBN architecture is its weight,w, which an RBM learns, defines both \(p(v|h, w)\)  and the prior distribution over hidden layers vector \(p(h|w)\). The probability of generating a visible vector 
\begin{equation}
p(v) = \sum_{h} p(h \mid w)\, p(v \mid h, w)
\end{equation}

\subsubsection{BiLSTM}
The BiLSTM model efficiently captures temporal dependencies in sequential data, thereby improving input representation by accurately tracking the input direction using forward and backward modules. Each is constructed with a bidirectional LSTM featuring distinct architectures \cite{m4}. These modules share a common output layer, allowing each unit to access comprehensive information from both past and future inputs simultaneously. Each operation in the model is associated with a weight \(w\). During Bi-LSTM operation, data flows through both forward and backward hidden states, yielding a hidden-layer output that incorporates bidirectional temporal processing \cite{m5}

\subsubsection{BIDIRECTIONAL ENCODER REPRESENTATIONS FROM TRANSFORMERS (BERT)}
\begin{figure}[htbp]
\centering
\includegraphics[width=\textwidth]{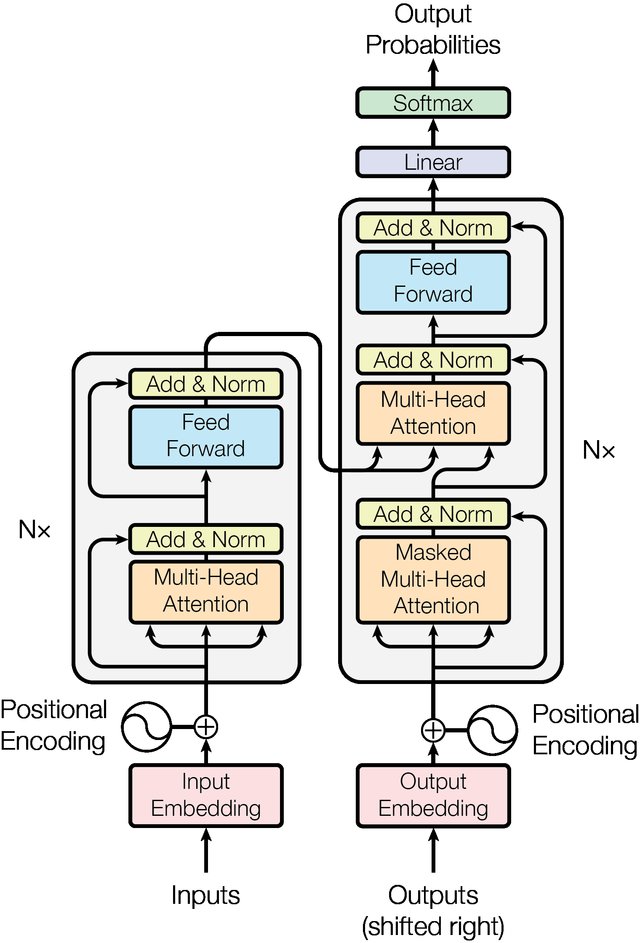}
\caption{Transnformer based model architecture with encoder and decoder \cite{m21}}\label{transformer}
\end{figure}

BERT is a transformer-based language model designed to learn deep word-level semantics by analyzing context. In contrast to other language models that process text sequentially from left to right, BERT's bidirectionality enables it to capture more subtle syntactic and semantic dependencies by simultaneously considering the entire sentence or passage. Its global perspective allows a model to capture subtle linguistic phenomena, such as polysemy (multiword meaning) and long-distance dependencies \cite{m6}. It pretrains BERT on large amounts of text using two general approaches. The first is masked language modeling, in which words in a sentence are randomly masked, and the Model is prompted to predict them from the context provided by the exposed words alone \cite{m7}. The architecture of the transformer based model are given in \ref{transformer}. The mask language model equation
\begin{equation}
 \mathcal{L}_{\text{MLM}} = -\sum_{i \in M} \log \frac{\exp(\mathbf{w}_{x_i}^\top \mathbf{h}_i)}{\sum_{j=1}^{V} \exp(\mathbf{w}_j^\top \mathbf{h}_i)}
\end{equation}
Here, \(M\) is masked token position, \({h}_i\), hidden state at ith position, and \({w}_j\) is the weight.
This leads BERT to learn how words interact with one another across diverse contexts. The second task, Next Sentence Prediction, trains the model to predict whether two given sentences are logically ordered, in the sense that it learns semantic relations between sentences, which is crucial for sentence-pair tasks. This pretraining makes BERT an extremely context-sensitive word- and sentence-embedding generator and, when fine-tuned on downstream natural language processing tasks such as NER, question answering, sentiment analysis, and text classification, performs exceptionally well. By incorporating more contextual information than previous models, BERT significantly improved the state of the art across a wide range of NLP tasks \cite{m8}.

\subsubsection{XLNET}  
XLNet is a transformer-based model that introduces a novel permutation-based training. The model is trained on each permutation of the input sequence rather than on a predetermined order. This method enables efficient management of bidirectional context without using masked tokens or artificial token masking for specific tokens \cite{m9}. XLNet adopts the Transformer-XL architecture and employs a recurrence mechanism that extends the context window beyond fixed-sized segments, thereby efficiently modeling long-range text dependencies. This makes XLNet particularly well-suited to learning and processing longer text, where it is desirable to capture how distant words relate to one another. By predicting tokens in random order, XLNet can discover patterns in higher-level dependencies and model language more effectively \cite{m10}.
\subsection{performance Metrics} 
To compare the performance of the proposed system, four key measures of performance accuracy, precision, recall, and F1-score were used to assess the proposed model. Accuracy is the proportion of correctly predicted values among all instances. It quantifies the model's overall accuracy by calculating the ratio of correctly labeled sensitive and non-sensitive entities to the total number of predictions. While helpful, it is misleading in biased datasets where sensitive data is less frequent. Precision is the proportion of true positive (TP) predictions among all positive predictions. Precision is used to determine the number of sensitive cases identified, thereby reducing false positives (FPs) and unnecessary encryption, which can affect performance and storage. Recall examines how well the model detects all the present sensitive data. It is significant that missing a positive instance, such as a false negative (FN), is more costly than misclassifying a negative instance as positive (false positive, FP). A high recall rate ensures that sensitive information is not left behind, reducing the risk of data breaches. Since precision and recall are in trade-off, the F1-score is the harmonic mean of the two, providing a balanced measure of the model's accuracy in identifying sensitive \cite{m3}. The equations for these metrics are given below:
\begin{equation}
\text{Accuracy} = \frac{TP + TN}{TP + TN + FP + FN}
\end{equation}
\begin{equation}
\text{Precision} = \frac{TP}{TP + FP}
\end{equation}
\begin{equation}
\text{Recall} = \frac{TP}{TP + FN}
\end{equation}
\begin{equation}
\text{F1\text{-}score} = 2 \times \frac{\text{Precision} \times \text{Recall}}{\text{Precision} + \text{Recall}}
\end{equation}

\section{Result and Discussion}
We have used five algorithms, such as BiLSTM, BERT, XLNet, LSTM, and DBN, with varied epoch numbers to measure the effectiveness of our proposed model structure of the transformer, which is given in Table \ref{tab:model_vertical}. We have compared results across two datasets to assess the efficacy of this combined grammar-based and contextual-feature approach in diverse domains, with respect to robustness and adaptability. Whether our model can learn complex patterns across diverse environments and text types. Additionally, a comparison of model performance with and without the grammar feature was conducted to assess its contribution to the overall effectiveness of the proposed method.

\subsection{Classification dataset performance}
\begin{figure}[htbp]
\centering
\includegraphics[width=\textwidth]{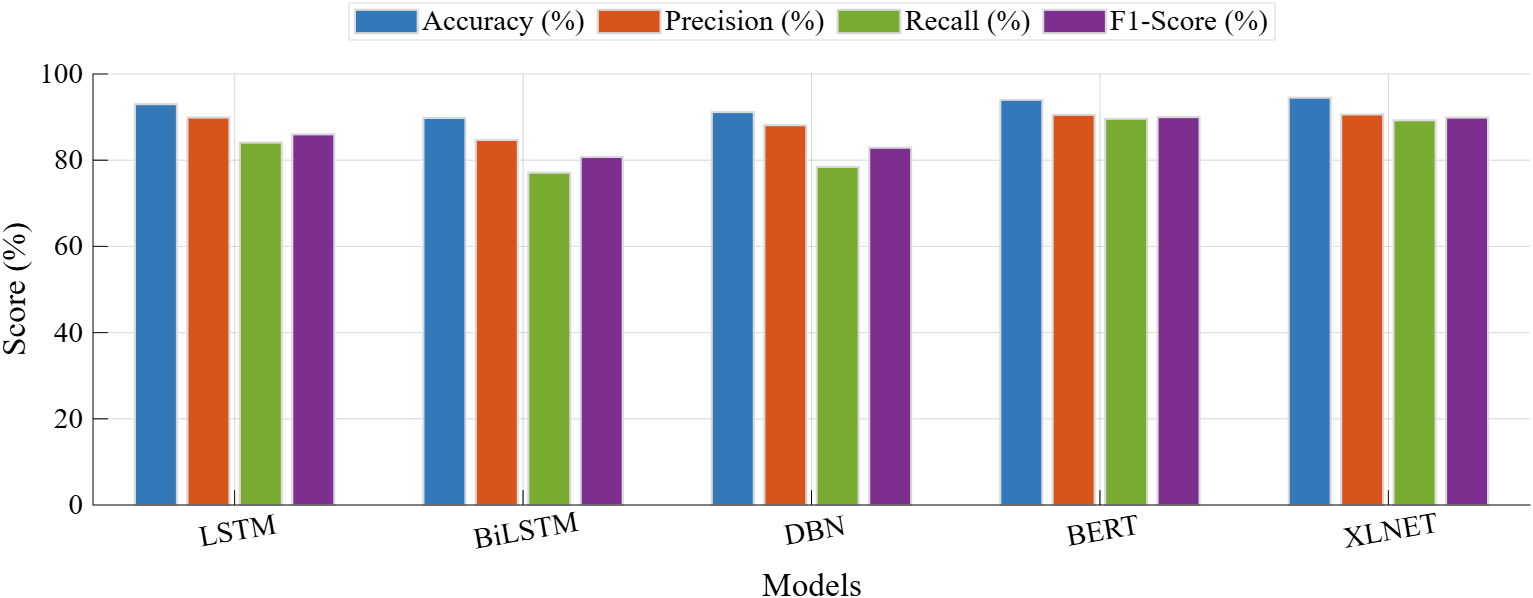}
\caption{Classification model performance without grammar features}\label{kbs_classfication_without_grammar_performance_bar}
\end{figure}

The table \ref{kbs_classfication_without_grammar_performance} presents the baseline performance of five algorithms, namely, LSTM, BiLSTM, DBN, BERT, and XLNet, without the proposed grammar-based features of the classification dataset. We used only contextual feature vectors, BERT tokenization, word embeddings, and TF-IDF, and compared model performance. Among the five algorithms, the BERT model performed best, achieving an accuracy of 93\%, with precision and recall of 90\%, and only the F1 score was below 90\%. XLNet model performance is also comparable to that of BERT. Among three deep learning algorithms, DBN, LSTM, and BiLSTM, LSTM has the highest accuracy of 93\%. The LSTM's recall value is very low, at 84\%, indicating that it missed many actual values. On the other hand, BiLSTM and DBN recall values are below 80\%. The BiLSTM demonstrates the lowest performance among the models. \ref{kbs_classfication_without_grammar_performance_bar}.

\begin{table}[tbp]
\centering
\caption{Performance Metrics of Different Algorithms without grammar feature}
\begin{tabular}{lcccc}
\toprule
\textbf{Model} & \textbf{Accuracy (\%)} & \textbf{Precision (\%)} & \textbf{Recall (\%)} & \textbf{F1-Score (\%)} \\
\midrule
LSTM    & 93.00  & 89.90  & 84.10  & 86.00  \\
BiLSTM  & 89.80  & 84.70  & 77.11  & 80.74  \\
DBN     & 91.15  & 88.12  & 78.43  & 82.90  \\
BERT    & 94.00  & 90.52  & 89.59  & 90.00  \\
XLNET   & 94.50  & 90.60  & 89.25  & 89.90  \\
\bottomrule
\end{tabular}
\label{kbs_classfication_without_grammar_performance}
\end{table}

\begin{figure}[htbp]
\centering
\includegraphics[width=\textwidth]{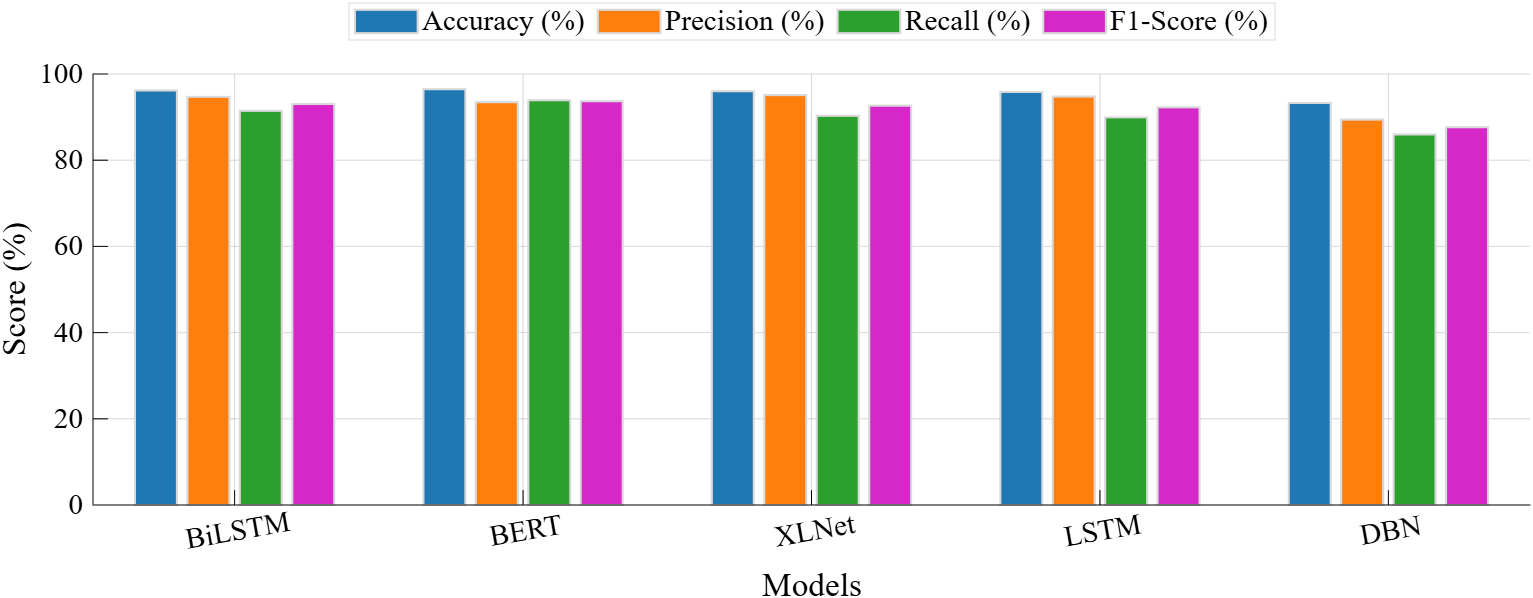}
\caption{Classification model performance with grammar features}\label{kbs_classfication_with_grammar_performance_bar}
\end{figure}

The table \ref{kbs_classfication_with_grammar_performance}. This table shows the performance classification dataset with the proposed feature fusion model that enhanced the model's overall performance by 2-7\%. BiLSTM performance increased significantly with an accuracy of 96.19\%, a precision of 94.69\%, a recall of 91.45\%, and an f1 score of 93\%. On the contrary, the performance metrics of the DBN algorithm are the lowest among the others, with a recall of 85\%; however, it showed progress compared with the baseline model. Additionally, the performance of BERT and XLNet is similar, as they are both transformer models, achieving an accuracy of 96\%. These comparisons highlight that fused feature vectors provide complementary structural information that significantly enhances the representational quality of sentence embeddings, leading to improved classification performance across diverse models \ref{kbs_classfication_with_grammar_performance_bar}.

\begin{table}[tbp]
\centering
\caption{Performance Metrics of Different Models with grammar-based features}
\begin{tabular}{lcccc}
\toprule
\textbf{Model} & \textbf{Accuracy (\%)} & \textbf{Precision (\%)} & \textbf{Recall (\%)} & \textbf{F1-Score (\%)} \\
\midrule
BiLSTM  & 96.19 & 94.69 & 91.45 & 93.04 \\
BERT    & 96.48 & 93.46 & 93.91 & 93.69 \\
XLNet   & 96.01 & 95.11 & 90.29 & 92.64 \\
LSTM    & 95.82 & 94.76 & 89.94 & 92.28 \\
DBN     & 93.27 & 89.43 & 85.97 & 87.66 \\
\bottomrule
\end{tabular}
\label{kbs_classfication_with_grammar_performance}
\end{table}

\subsection{NER dataset performance}
\begin{figure}[tbp]
\centering
\includegraphics[width=\textwidth]{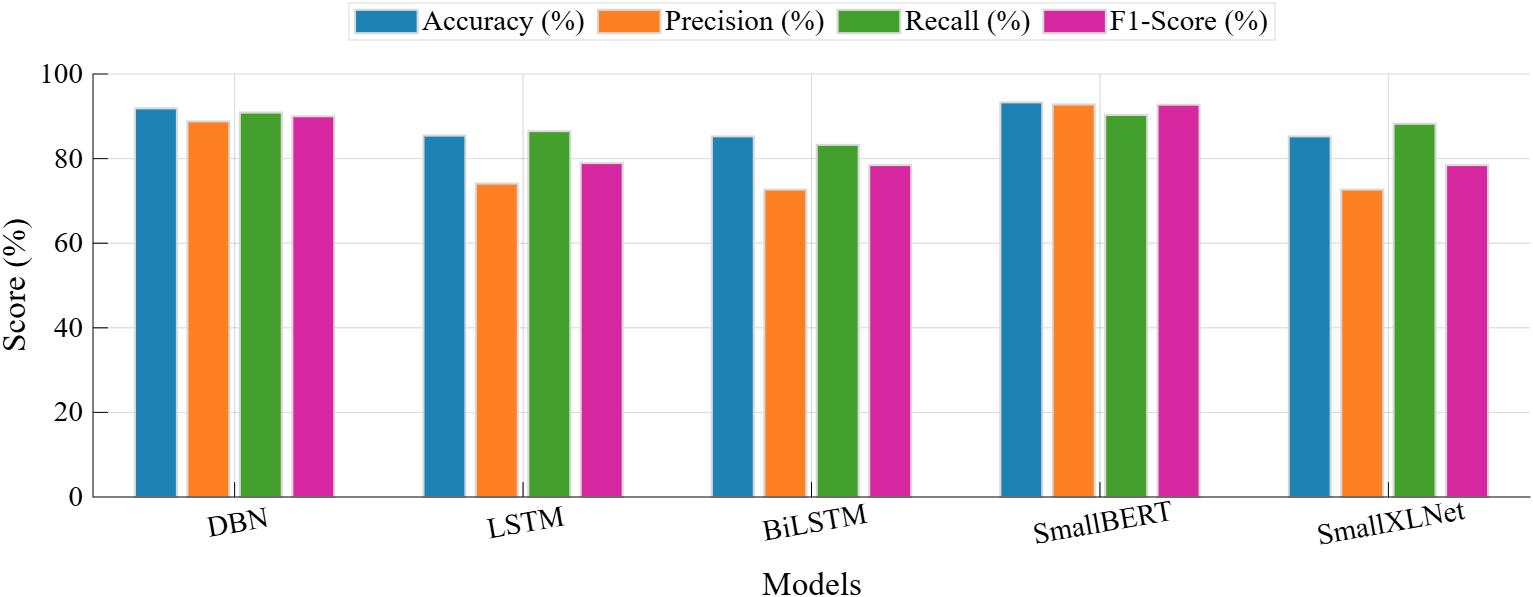}
\caption{NER model performance without grammar features}\label{kbs_NER_without_grammar_performance_bar}
\end{figure}

The performance of the NER model without a grammar-based feature vector is demonstrated in Table \ref{kbs_NER_without_grammar_performance}. According to the table, the BERT model performed best, achieving an accuracy of 93.29\%, precision of 92.8\%, recall of 90.32\%, and F1 score of 92.73\%. Bert is the only model whose performance across all metrics exceeds 90\%. Except for the precision value (88.80\%). The DBN algorithm's precision value is 88.8\%. Both LSTM (85.45\% accuracy) and BiLSTM (85.25\% accuracy) show similar moderate performance with notably lower precision around 72-74\%, indicating they produce more false positives. In this NER dataset, XLNet performed worst with the lowest accuracy of 84.53\% and precision of 71.61\% depicts in Figure \ref{kbs_NER_without_grammar_performance_bar}

\begin{table}[tbp]
\centering
\caption{Performance Metrics of NER models without grammatical feature}
\begin{tabular}{lcccc}
\toprule
\textbf{Model} & \textbf{Accuracy (\%)} & \textbf{Precision (\%)} & \textbf{Recall (\%)} & \textbf{F1-Score (\%)} \\
\midrule
DBN & 91.88 & 88.80 & 90.88 & 90.01 \\
LSTM & 85.45 & 74.10 & 86.5 & 78.93 \\
BiLSTM & 85.25 & 72.67 & 83.25 & 78.46 \\
BERT & 93.29 & 92.80 & 90.32 & 92.73 \\
XLNet & 84.53 & 71.61 & 88.26 & 79.43 \\
\bottomrule
\end{tabular}
\label{kbs_NER_without_grammar_performance}
\end{table}

\begin{figure}[tbp]
\centering
\includegraphics[width=\textwidth]{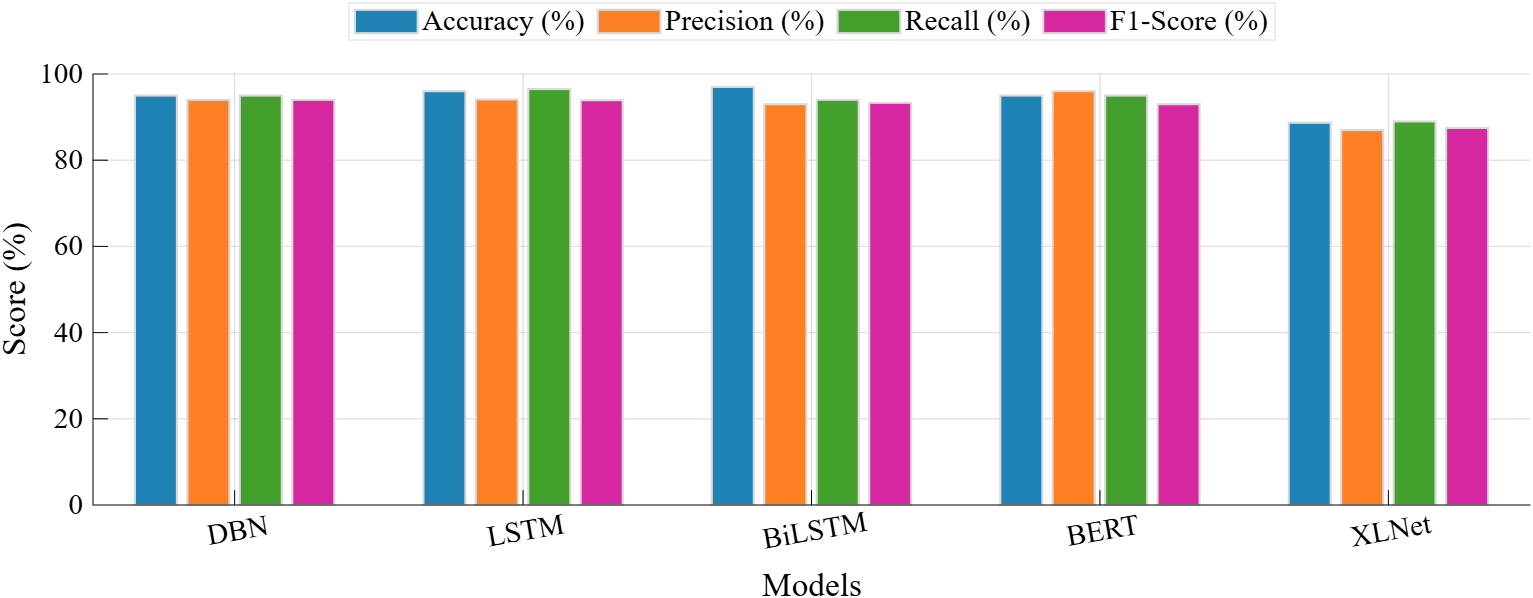}
\caption{NER model performance without grammar features}\label{kbs_NER_with_grammar_performance_bar}
\end{figure}

According to the table \ref{kbs_NER_with_grammar_performance}, the performance of all algorithms increased dramatically, except XLNet. The accuracy of the DBN algorithm increased by almost 4\% compared with feature training without grammar rules. Moreover, precision and recall increased by approximately 5\%, from 88\% to 94\%. The accuracy of the LSTM and BiLSTM models increased significantly, from 85\% to 97\% for the BiLSTM model. When the BiLSTM model was trained without a combined feature vector, the precision and F1 scores of these two algorithms were below 80\%; however, with our grammar-based feature model, they exceeded 90\%, as shown in Figure \ref{kbs_NER_with_grammar_performance_bar}. Thus, the NER model performs better when feature vectors are combined with contextual and linguistic information.
\begin{table}[h]
\centering
\caption{Performance Metrics of NER models with grammatical feature}
\label{kbs_NER_with_grammar_performance}
\begin{tabular}{lcccc}
\toprule
\textbf{Model} & \textbf{Accuracy (\%)} & \textbf{Precision (\%)} & \textbf{Recall (\%)} & \textbf{F1-Score (\%)} \\
\midrule
DBN & 95 & 94 & 95 & 94 \\
LSTM & 96 & 94.1 & 96.5 & 93.93 \\
BiLSTM & 97 & 93 & 94 & 93.3 \\
BERT & 95 & 96 & 95 & 93 \\
XLNet & 88.67 & 87 & 89 & 87.48 \\
\bottomrule
\end{tabular}
\end{table}
The observed performance gains, achieved without transformer fine-tuning or additional syntactic encoders, indicate that explicit grammatical inductive biases can substitute for architectural complexity, enabling lightweight models to approach or exceed the performance of fully fine-tuned transformers in cross-domain settings.

\section{discussion}

\begin{sidewaystable}[ht]
\centering
\caption{Model Architecture Comparison}
\label{tab:model_vertical}
\footnotesize
\begin{tabular}{%
p{2.2cm} 
p{3.0cm} 
p{2.0cm} 
p{2.0cm} 
p{2.0cm} 
p{2.2cm} 
p{2.5cm} 
p{2.8cm} 
}
\toprule
\textbf{Model} & \textbf{Backbone Used} & \textbf{Layers} & \textbf{Hidden Size} & \textbf{Attention Heads} & \textbf{Total Parameters} & \textbf{Trainable Parameters} & \textbf{Approx. Model Size (FP32)} \\
\midrule
LSTM & Trainable Embeddings & 5 layers  & 64 & N/A & ~1.33M & ~1.33M & ~5.32 MB \\
BiLSTM & Trainable Embeddings & 5 layers & 48×2 & N/A & ~1.38M & ~1.38M & ~5.52 MB \\
BERT (proposed model) & Frozen MiniLM-L6-v2 & 3 layers & 256→128→64 & N/A & ~157K & ~157K & ~0.63 MB \\
XLNet (proposed model) & Frozen MiniLM-L6-v2 & 3 layers & 192→96→48 & N/A & ~112K & ~112K & ~0.45 MB \\
DBN & PCA-TF-IDF (128-dim) & 2 layers & 256→64 & N/A & ~49K & ~49K & ~0.20 MB \\
BERT Backbone & MiniLM-L6-v2 & 6 Transformer & 384 & 12 & ~22.7M & 0 (frozen) & ~90.8 MB \\
Full BERT-base & BERT-base & 12 Transformer & 768 & 12 & ~110M & ~110M & ~440 MB \\
Full BERT-large & BERT-large & 24 Transformer & 1024 & 16 & ~340M & ~340M & ~1.36 GB \\
Full XLNet-base & XLNet-base & 12 Transformer & 768 & 12 & ~110M & ~110M & ~440 MB \\
Full XLNet-large & XLNet-large & 24 Transformer & 1024 & 16 & ~340M & ~340M & ~1.36 GB\\
\bottomrule
\end{tabular}

\end{sidewaystable}

This study introduces a novel hybrid framework that integrates grammar-aware linguistic features with transformer-based embeddings to improve robustness in domain-divergent linguistic settings text classification. By combining syntactic structure with deep contextual semantics, the proposed approach effectively captures domain-invariant patterns and enhances generalization to unseen domains. Our proposed grammar-based feature model has 11 attributes, providing extensive grammatical information about the sentence. Details are given in the Table \ref{tab:feature_groups_detailed}. Our framework first extracts structured grammatical features and then fuses them with Transformer-generated embeddings to train a robust model. The experimental results demonstrate that integrating grammar-aware feature extraction significantly enhances cross-domain text classification performance. The binary classification dataset performance without grammar features was better, especially for transformer-based models, which can learn more effectively through their attention mechanisms. However, the combined grammar and text-based feature vector increased overall performance from 2\% to 7\% in the binary classification dataset. In the NER dataset, our proposed grammar-based features performed best, as entity identification also requires grammar rules. Because the NER model requires proper grammatical structure to detect entities accurately, our proposed feature vector combines grammatical and contextual information and could be transformative for textual models. From Table \ref{tab:model_vertical}, BERT-base and XLNet-base, which are standard 12-layer architectures, take almost 440 MB to train, whereas BERT-large and XLNet-large, with 24-layer versions and larger hidden dimensions, are 1.36GB in size and have 340 million parameters each. On the contrary, our proposed model used only a frozen BERT model to learn only semantic information. 
Concatenating them with frozen embeddings adds very few extra parameters compared to training a full transformer. Transformers learn general contextual embeddings, but some domain-specific patterns are subtle. Grammar features provide structured clues that the model might otherwise take longer to learn from data. Grammar features are symbolic and don’t depend on patterns in large datasets. When you train on one domain and test on another, grammar features help maintain performance. and Embeddings capture semantic meaning, grammar features capture syntactic structure. Concatenating both more informative representations without adding trainable transformer parameters. Grammar features provide prior knowledge, reducing overfitting and boosting accuracy with fewer samples. Therefore, our proposed model is not only lightweight but also maintains performance through smart concatenation with contextual features and grammar features.
\section{Conclusion}
This paper proposes an extensive grammar-based feature-extraction algorithm that combines a deep neural network and a transformer-based model. Whereas transformer-based models’ feature vectors capture semantic and syntactic information, often by pre-training on large-scale text corpora, they usually fail to capture in-depth grammatical structures and perform better on domain-specific corpora. To verify the effectiveness of the proposed model, a series of comparative experiments is carried out. To assess the model's robustness, our proposed model uses two datasets: classification datasets and an NER dataset. The experimental results indicate that the proposed feature extraction model effectively extracts features across domains, with an average increase of 2\%-15\% in accuracy. The approach systematically leverages grammar-aware features alongside transformer embeddings for robust domain-divergent text classification in low-resource settings, leading to low-resource model design. Our lightweight framework combines frozen transformer representations with explicit grammar-based features, drastically reducing computational overhead without sacrificing accuracy. This synergy enables efficient learning and improved generalization compared to fully fine-tuned models.

The main limitation of our proposed system is that grammatical structures vary substantially across domains (e.g., social media text versus formal writing). We tested only our proposed system on the email classification model and the NER model. Testing on a diverse domain would generate more accurate reasoning for our system. Combining sparse grammar vectors with dense contextual embeddings requires specialized fusion strategies; however, resource limitations in our model led us to use a concatenation-based data fusion method. 

In the future, we can integrate a tree-based encoder to learn syntactic structures directly from data and evaluate the model across a wider range of domains in a multilingual setting. The proposed framework offers a scalable and reliable approach to cross-domain text classification, contributing meaningful insights to the development of domain-agnostic natural language processing systems.
\bibliographystyle{IEEEtran}
\bibliography{reference.bib}

@article{i1,
  title={Natural language processing},
  author={Chowdhary, KR1442},
  journal={Fundamentals of artificial intelligence},
  pages={603--649},
  year={2020},
  publisher={Springer}
}

@article{i2,
  title={Natural language processing: an introduction},
  author={Nadkarni, Prakash M and Ohno-Machado, Lucila and Chapman, Wendy W},
  journal={Journal of the American Medical Informatics Association},
  volume={18},
  number={5},
  pages={544--551},
  year={2011},
  publisher={BMJ Group BMA House, Tavistock Square, London, WC1H 9JR}
}

@techreport {i3,
  author       = {Megha Gupta},
  title        = {Global Natural Language Processing (NLP) Market – Industry Trends and Forecast to 2030},
  institution  = {Data Bridge Market Research},
  year         = {2023},
  month        = sep,
  url          = {https://www.databridgemarketresearch.com/reports/global-natural-language-processing-nlp-market},
  note         = {Market Research Report, 350 pages, 220 tables, 60 figures}
}

@article{i4,
  title={Advances in natural language processing},
  author={Hirschberg, Julia and Manning, Christopher D},
  journal={Science},
  volume={349},
  number={6245},
  pages={261--266},
  year={2015},
  publisher={American Association for the Advancement of Science}
}

@article{i5,
  title={A systematic review of natural language processing for classification tasks in the field of incident reporting and adverse event analysis},
  author={Young, Ian James Bruce and Luz, Saturnino and Lone, Nazir},
  journal={International journal of medical informatics},
  volume={132},
  pages={103971},
  year={2019},
  publisher={Elsevier}
}

@article{i6,
  title={Recent advancements and challenges of NLP-based sentiment analysis: A state-of-the-art review},
  author={Jim, Jamin Rahman and Talukder, Md Apon Riaz and Malakar, Partha and Kabir, Md Mohsin and Nur, Kamruddin and Mridha, Mohammed Firoz},
  journal={Natural Language Processing Journal},
  volume={6},
  pages={100059},
  year={2024},
  publisher={Elsevier}
}

@inproceedings{i7,
  title={Evaluating natural language understanding services for conversational question answering systems},
  author={Braun, Daniel and Mendez, Adrian Hernandez and Matthes, Florian and Langen, Manfred},
  booktitle={Proceedings of the 18th annual SIGdial meeting on discourse and dialogue},
  pages={174--185},
  year={2017}
}

@inproceedings{i8,
  title={Natural language processing and its applications in machine translation: a diachronic review},
  author={Jiang, Kai and Lu, Xi},
  booktitle={2020 IEEE 3rd international conference of safe production and informatization (IICSPI)},
  pages={210--214},
  year={2020},
  organization={IEEE}
}

@inproceedings{i9,
  title={Natural language processing (NLP) based text summarization-a survey},
  author={Awasthi, Ishitva and Gupta, Kuntal and Bhogal, Prabjot Singh and Anand, Sahejpreet Singh and Soni, Piyush Kumar},
  booktitle={2021 6th International Conference on Inventive Computation Technologies (ICICT)},
  pages={1310--1317},
  year={2021},
  organization={IEEE}
}

@article{i10,
  title={Natural language processing (NLP) in management research: A literature review},
  author={Kang, Yue and Cai, Zhao and Tan, Chee-Wee and Huang, Qian and Liu, Hefu},
  journal={Journal of Management Analytics},
  volume={7},
  number={2},
  pages={139--172},
  year={2020},
  publisher={Taylor \& Francis}
}

@inproceedings{i12,
  title={An empirical investigation of statistical significance in NLP},
  author={Berg-Kirkpatrick, Taylor and Burkett, David and Klein, Dan},
  booktitle={Proceedings of the 2012 joint conference on empirical methods in natural language processing and computational natural language learning},
  pages={995--1005},
  year={2012}
}

@article{i13,
  title={Understanding the difficulty of training transformers},
  author={Liu, Liyuan and Liu, Xiaodong and Gao, Jianfeng and Chen, Weizhu and Han, Jiawei},
  journal={arXiv preprint arXiv:2004.08249},
  year={2020}
}

@article{i14,
  title={A conceptual basis for feature engineering},
  author={Turner, C Reid and Fuggetta, Alfonso and Lavazza, Luigi and Wolf, Alexander L},
  journal={Journal of Systems and Software},
  volume={49},
  number={1},
  pages={3--15},
  year={1999},
  publisher={Elsevier}
}

@inproceedings{i15,
  title={Feature engineering for text classification},
  author={Scott, Sam and Matwin, Stan},
  booktitle={ICML},
  volume={99},
  pages={379--388},
  year={1999}
}

@article{i16,
  title={Feature extraction and analysis of natural language processing for deep learning English language},
  author={Wang, Dongyang and Su, Junli and Yu, Hongbin},
  journal={IEEE Access},
  volume={8},
  pages={46335--46345},
  year={2020},
  publisher={IEEE}
}

@article{i17,
  title={An ensemble scheme based on language function analysis and feature engineering for text genre classification},
  author={Onan, Aytu{\u{g}}},
  journal={Journal of Information Science},
  volume={44},
  number={1},
  pages={28--47},
  year={2018},
  publisher={SAGE Publications Sage UK: London, England}
}

@inproceedings{i18,
  title={Text classification using different feature extraction approaches},
  author={Dzisevi{\v{c}}, Robert and {\v{S}}e{\v{s}}ok, Dmitrij},
  booktitle={2019 Open Conference of Electrical, Electronic and Information Sciences (eStream)},
  pages={1--4},
  year={2019},
  organization={IEEE}
}

@article{i19,
  title={A detailed review on word embedding techniques with emphasis on word2vec},
  author={Johnson, S Joshua and Murty, M Ramakrishna and Navakanth, I},
  journal={Multimedia Tools and Applications},
  volume={83},
  number={13},
  pages={37979--38007},
  year={2024},
  publisher={Springer}
}

@article{i20,
  title={Glove prototype for feature extraction applied to learning by demonstration purposes},
  author={Cerqueira, Tiago and Ribeiro, Francisco M and Pinto, V{\'\i}tor H and Lima, Jos{\'e} and Gon{\c{c}}alves, Gil},
  journal={Applied Sciences},
  volume={12},
  number={21},
  pages={10752},
  year={2022},
  publisher={MDPI}
}

@inproceedings{i21,
  title={Fastext: Efficient unconstrained scene text detector},
  author={Busta, Michal and Neumann, Lukas and Matas, Jiri},
  booktitle={Proceedings of the IEEE international conference on computer vision},
  pages={1206--1214},
  year={2015}
}

@article{i22,
  title={Feature extraction and analysis of natural language processing for deep learning English language},
  author={Wang, Dongyang and Su, Junli and Yu, Hongbin},
  journal={IEEE Access},
  volume={8},
  pages={46335--46345},
  year={2020},
  publisher={IEEE}
}

@article{r1,
  title={Enhancing Text Classification Through Grammar-Based Feature Engineering and Learning Models},
  author={Mohasseb, Alaa and Kanavos, Andreas and Amer, Eslam},
  journal={Information},
  volume={16},
  number={6},
  pages={424},
  year={2025},
  publisher={MDPI}
}

@article{r2,
  title={Firefly algorithm based feature selection for Arabic text classification},
  author={Marie-Sainte, Souad Larabi and Alalyani, Nada},
  journal={Journal of King Saud University-Computer and Information Sciences},
  volume={32},
  number={3},
  pages={320--328},
  year={2020},
  publisher={Elsevier}
}

@inproceedings{r3,
  title={Grammar detection for sentiment analysis through improved viterbi algorithm},
  author={Chavali, Surya Teja and Kandavalli, Charan Tej and others},
  booktitle={2022 International Conference on Advances in Computing, Communication and Applied Informatics (ACCAI)},
  pages={1--6},
  year={2022},
  organization={IEEE}
}

@inproceedings{r4,
  title={A novel part of speech tagging framework for nlp based business process management},
  author={Han, Xue and Dang, Yabin and Mei, Lijun and Wang, Yanfei and Li, Shaochun and Zhou, Xin},
  booktitle={2019 IEEE International Conference on Web Services (ICWS)},
  pages={383--387},
  year={2019},
  organization={IEEE}
}

@article{r5,
  title={Part-of-Speech tagging enhancement to natural language processing for Thai wh-question classification with deep learning},
  author={Chotirat, Saranlita and Meesad, Phayung},
  journal={Heliyon},
  volume={7},
  number={10},
  year={2021},
  publisher={Elsevier}
}

@article{r6,
  title={Part-of-speech tagging with rule-based data preprocessing and transformer},
  author={Li, Hongwei and Mao, Hongyan and Wang, Jingzi},
  journal={Electronics},
  volume={11},
  number={1},
  pages={56},
  year={2021},
  publisher={MDPI}
}

@article{r7,
  title={Evaluation of feature selection methods for text classification with small datasets using multiple criteria decision-making methods},
  author={Kou, Gang and Yang, Pei and Peng, Yi and Xiao, Feng and Chen, Yang and Alsaadi, Fawaz E},
  journal={Applied Soft Computing},
  volume={86},
  pages={105836},
  year={2020},
  publisher={Elsevier}
}

@article{r8,
  title={Enhancing text classification using hybrid embeddings and advanced machine learning techniques},
  author={Chaudhary, Abhishek and Jain, Shikha},
  journal={International Journal of System Assurance Engineering and Management},
  pages={1--16},
  year={2025},
  publisher={Springer}
}

@article{r9,
  title={Embedding generation for text classification of Brazilian Portuguese user reviews: from bag-of-words to transformers},
  author={Souza, Frederico Dias and Filho, Jo{\~a}o Baptista de Oliveira e Souza},
  journal={Neural Computing and Applications},
  volume={35},
  number={13},
  pages={9393--9406},
  year={2023},
  publisher={Springer}
}

@article{r10,
  title={Short text understanding combining text conceptualization and transformer embedding},
  author={Li, Jun and Huang, Guimin and Chen, Jianheng and Wang, Yabing},
  journal={IEEE Access},
  volume={7},
  pages={122183--122191},
  year={2019},
  publisher={IEEE}
}

@incollection{m1,
  title={The groningen meaning bank},
  author={Bos, Johan and Basile, Valerio and Evang, Kilian and Venhuizen, Noortje J and Bjerva, Johannes},
  booktitle={Handbook of linguistic annotation},
  pages={463--496},
  year={2017},
  publisher={Springer}
}

@inproceedings{m2,
  title={The enron corpus: A new dataset for email classification research},
  author={Klimt, Bryan and Yang, Yiming},
  booktitle={European conference on machine learning},
  pages={217--226},
  year={2004},
  organization={Springer}
}

@article{m3,
  title={Toward a better parallel performance metric},
  author={Sun, Xian-He and Gustafson, John L},
  journal={Parallel Computing},
  volume={17},
  number={10-11},
  pages={1093--1109},
  year={1991},
  publisher={Elsevier}
}

@article{m4,
  title={A hybrid optimization algorithm using BiLSTM structure for sentiment analysis},
  author={Sangeetha, J and Kumaran, U},
  journal={Measurement: Sensors},
  volume={25},
  pages={100619},
  year={2023},
  publisher={Elsevier}
}

@article{m5,
  title={Reconstruction of structural long-term acceleration response based on BiLSTM networks},
  author={Lu, Yonghui and Tang, Liqun and Chen, Chengbin and Zhou, Licheng and Liu, Zejia and Liu, Yiping and Jiang, Zhenyu and Yang, Bao},
  journal={Engineering Structures},
  volume={285},
  pages={116000},
  year={2023},
  publisher={Elsevier}
}

@inproceedings{m6,
  title={What does BERT learn about the structure of language?},
  author={Jawahar, Ganesh and Sagot, Beno{\^\i}t and Seddah, Djam{\'e}},
  booktitle={ACL 2019-57th Annual Meeting of the Association for Computational Linguistics},
  year={2019}
}

@article{m7,
  title={BERT: a review of applications in natural language processing and understanding},
  author={Koroteev, Mikhail V},
  journal={arXiv preprint arXiv:2103.11943},
  year={2021}
}

@article{m8,
  title={What does bert look at? an analysis of bert's attention},
  author={Clark, Kevin and Khandelwal, Urvashi and Levy, Omer and Manning, Christopher D},
  journal={arXiv preprint arXiv:1906.04341},
  year={2019}
}

@article{m9,
  title={Xlnet: Generalized autoregressive pretraining for language understanding},
  author={Yang, Zhilin and Dai, Zihang and Yang, Yiming and Carbonell, Jaime and Salakhutdinov, Russ R and Le, Quoc V},
  journal={Advances in neural information processing systems},
  volume={32},
  year={2019}
}

@inproceedings{m10,
  title={Comparative analyses of bert, roberta, distilbert, and xlnet for text-based emotion recognition},
  author={Adoma, Acheampong Francisca and Henry, Nunoo-Mensah and Chen, Wenyu},
  booktitle={2020 17th international computer conference on wavelet active media technology and information processing (ICCWAMTIP)},
  pages={117--121},
  year={2020},
  organization={IEEE}
}

@incollection{m12,
  title={Syntactic parsing},
  author={Pickering, Martin J and Van Gompel, Roger PG},
  booktitle={Handbook of psycholinguistics},
  pages={455--503},
  year={2006},
  publisher={Elsevier}
}

@article{m13,
  title={Comparison of text preprocessing methods},
  author={Chai, Christine P},
  journal={Natural language engineering},
  volume={29},
  number={3},
  pages={509--553},
  year={2023},
  publisher={Cambridge University Press}
}

@incollection{m14,
  title={Natural language processing (NLP): An introduction: making sense of textual data},
  author={Egger, Roman and Gokce, Enes},
  booktitle={Applied data science in tourism: Interdisciplinary approaches, methodologies, and applications},
  pages={307--334},
  year={2022},
  publisher={Springer}
}

@article{m15,
  title={Feature extraction and analysis of natural language processing for deep learning English language},
  author={Wang, Dongyang and Su, Junli and Yu, Hongbin},
  journal={IEEE Access},
  volume={8},
  pages={46335--46345},
  year={2020},
  publisher={IEEE}
}

@article{m16,
  title={POS tagging approaches: A comparison},
  author={Kumawat, Deepika and Jain, Vinesh},
  journal={International Journal of Computer Applications},
  volume={118},
  number={6},
  year={2015},
  publisher={Foundation of Computer Science}
}

@article{m17,
  title={The role of lexicons in natural language processing},
  author={Guthrie, Louise and Pustejovsky, James and Wilks, Yorick and Slator, Brian M},
  journal={Communications of the ACM},
  volume={39},
  number={1},
  pages={63--72},
  year={1996},
  publisher={ACM New York, NY, USA}
}

@inproceedings{m18,
  title={Shallow NLP techniques for noun phrase extraction},
  author={Subhashini, R and Kumar, V Jawahar Senthil},
  booktitle={Trendz in information sciences \& computing (tisc2010)},
  pages={73--77},
  year={2010},
  organization={IEEE}
}

@article{m19,
  title={Natural language processing in support of decision-making: phrases and part-of-speech tagging},
  author={Losee, Robert M},
  journal={Information processing \& management},
  volume={37},
  number={6},
  pages={769--787},
  year={2001},
  publisher={Elsevier}
}

@inproceedings{m20,
  title={A study of parsing process on natural language processing in bahasa Indonesia},
  author={Sibarani, Elisa Margareth and Nadial, Mhd and Panggabean, Evy and Meryana, S},
  booktitle={2013 IEEE 16th International Conference on Computational Science and Engineering},
  pages={309--316},
  year={2013},
  organization={IEEE}
}

@inproceedings{m21,
  title={Tensor2tensor for neural machine translation},
  author={Vaswani, Ashish and Bengio, Samy and Brevdo, Eugene and Chollet, Francois and Gomez, Aidan and Gouws, Stephan and Jones, Llion and Kaiser, {\L}ukasz and Kalchbrenner, Nal and Parmar, Niki and others},
  booktitle={Proceedings of the 13th Conference of the Association for Machine Translation in the Americas (Volume 1: Research Track)},
  pages={193--199},
  year={2018}
}
\end{document}